\documentclass[letterpaper, 10 pt, conference]{ieeeconf}
\IEEEoverridecommandlockouts             
\usepackage{amsmath,amsfonts}
\usepackage{algorithmic}
\usepackage{algorithm}
\usepackage{array}
\usepackage{textcomp}
\usepackage{stfloats}
\usepackage{url}
\usepackage{verbatim}
\usepackage{graphicx}
\usepackage{cite}
\usepackage[export]{adjustbox}
\usepackage{hyperref}
\usepackage{xurl}
\usepackage[linesnumbered,ruled,vlined,algo2e]{algorithm2e}
\usepackage{amssymb}
\usepackage[ansinew]{inputenc} 
\usepackage{xcolor}
\usepackage{mathtools}
\usepackage{caption}
\usepackage{subcaption}
\usepackage{import}
\usepackage{multirow}
\usepackage{cite}
\usepackage{breqn}
\usepackage{mathrsfs}
\usepackage{acronym}
\usepackage{setspace}
\usepackage{bm}
\usepackage{stackengine}
\usepackage{stackengine}
\usepackage{needspace}
\usepackage{comment}
\usepackage{siunitx}
\usepackage{lipsum}

\usepackage{amsthm}
\usepackage{amssymb}
\usepackage{svg}
\usepackage{lipsum}
\usepackage{wasysym}
\usepackage{mathrsfs}
\usepackage[T1]{fontenc}

\theoremstyle{plain}

\theoremstyle{definition}

\newtheorem{problem}{Problem}

\theoremstyle{remark}
\newtheorem{remark}{Remark}

\begin{document}
\title{\LARGE \bf
  From Pixels to Shelf: An Integrated Robotic System for Autonomous Supermarket Stocking with a Mobile Manipulator
}

\author{Davide Peron$^1$, Victor Nan Fernandez-Ayala$^2$, Lukas Segelmark$^3$ 
\thanks{$^1$ Department of Information Engineering, University of Padova, 35122, Padova, Italy. Email: {\tt\small davide.peron.3@studenti.unipd.it}} 
\thanks{$^2$ Division of Decision
and Control Systems, School of Electrical Engineering and Computer
Science, KTH Royal Institute of Technology, 100 44, Stockholm, Sweden. Email: \tt\small vnfa@kth.se}
\thanks{$^3$ Animum AB, 114 28, Stockholm, Sweden. Email: {\tt\small lukas@animum.ai}}
}

\maketitle
\thispagestyle{empty}
\pagestyle{empty}

\newcommand{\until}[0]{\mathsf{U}}

\def\triangleq{\mathrel{\ensurestackMath{\stackon[1pt]{=}{\scriptstyle\Delta}}}}

\maketitle

\SetKwFunction{SAMSegment}{SAM2\_segmentation}
\SetKwFunction{SAMtrack}{SAM2\_Tracking}
\SetKwFunction{yologrocery}{YOLO\_grocery}
\SetKwFunction{selectprodyolo}{YOLO\_Prod}

\begin{abstract}
Autonomous stocking in retail environments, particularly supermarkets, presents challenges due to dynamic human interactions, constrained spaces, and diverse product geometries. This paper introduces an efficient modular robotic system for autonomous shelf stocking, integrating commercially available hardware with a scalable algorithmic architecture. A major contribution of this work is the system integration of off-the-shelf hardware and ROS2-based perception, planning, and control into a single deployable platform for retail environments. Our solution leverages Behavior Trees (BTs) for task planning, fine-tuned vision models for object detection, and a two-step Model Predictive Control (MPC) framework for precise shelf navigation using ArUco markers. Laboratory experiments replicating realistic supermarket conditions demonstrate reliable performance, achieving over 98\% success in pick-and-place operations across a total of more than 700 stocking events. However, our comparative benchmarks indicate that the performance and cost-effectiveness of current autonomous systems remain inferior to that of human workers, which we use to highlight key improvement areas and quantify the progress still required before widespread commercial deployment can realistically be achieved.
\end{abstract}

\section{Introduction} \label{sec:introduction}
In today's global economy, industries face mounting pressure from labor shortages, rising operational costs, and growing demands for efficiency. Autonomous mobile robots (AMRs) have emerged as a promising solution to address these challenges. AMRs are increasingly being deployed across various sectors such as manufacturing~\cite{manufacturing}, and  agriculture~\cite{ICRA25_recurringLTL}, where they perform repetitive and physically demanding tasks to improve productivity and safety standards.

The retail sector, specifically the grocery market, presents attractive opportunities for robotic automation. In the United States alone, more than 1.5 million workers are currently employed as stockers and order fillers across food, beverage, and general merchandise retailers~\cite{bls_stockers_2024}. Shelf stocking and product arrangement operations consume over 30\% of store attendants' working hours~\cite{guney2019forecasting}, representing a significant portion of the operational cost. AMRs have already shown considerable success in structured warehouse scenarios, such as Amazon's fulfillment centers~\cite{amazon}. However, comparable adoption in retail environments remains limited due to significant technical challenges, high costs, and complexity posed by dynamic and shared human-robot environments~\cite{airlab_paper}.

Recent advancements in retail robotics focus mainly on inventory tracking and shelf scanning, order picking, and shelf restocking~\cite{robotics11050104, sasono2025planning}. Bossa Nova's robots used by Walmart~\cite{bossa_nova} have shown effectiveness in inventory management through shelf monitoring tasks without manipulation capabilities. More recent end-to-end mobile manipulation has been explored by Spahn et al.~\cite{airlab_paper}, Huang et al.~\cite{shopper},  and Wu et al.~\cite{in_the_wild}, which demonstrate autonomous item picking in store-like aisles but rely on higher-end hardware, more constrained assumptions than ours, and do not explicitly address shelf stocking challenges or the "last-meter" navigation problem~\cite{costanzo}. Other work targets submodules rather than complete pipelines: vacuum-based grasping for densely packed shelves~\cite{s24206687}, which offers a complementary view to our gripper setup and a promising avenue for future improvements, bimanual replenishment in highly controlled scenes \cite{robotics11050104}, and perception-only planogram compliance and product localization~\cite{PIETRINI2024124635, s25175309}.

To address these limitations, we propose an integrated, cost-effective robotic system for autonomous shelf stocking, tailored to retail supermarket environments. Our system targets cost-constrained deployment by using a commercially available and affordable mobile manipulator: the Hello Robot Stretch 3~\cite{stretch}, combined with a structured and scalable algorithmic architecture. By integrating the Robot Operating System 2 (ROS2) framework, Behavior Trees (BTs) for planning, and advanced perception and control methodologies, including YOLO-based object detection~\cite{yolo} and SAM2 for segmentation and tracking~\cite{sam2}, we enable reliable and adaptable operations. Specifically, our approach addresses critical retail-specific issues such as precise product manipulation, accurate localization, and robust navigation, using ArUco markers~\cite{aruco}, a Kalman Filter (KF)~\cite{Kalman_1960}, and a two step Model Predictive Control (MPC)~\cite{mpc} for integrated base and camera motion, mitigating the challenges posed by conventional sensor limitations in "last-meter" navigation.

This work makes three contributions. First, we
present a robust system integration of off-the-shelf hardware with a ROS2 stack, validating the combined effectiveness of BTs task planning, Nav2~\cite{nav2}, ArUco-KF localization, and a two-step base-head MPC for "last-meter" shelf navigation, together with YOLO and SAM2 perception and planogram-aware Inverse Kinematics (IK) controller for reliable retail manipulation. The key modules and simulator are released as open source \cite{our_docking,our_nav2, ros_bridge, our_simulator}. Second, compared with prior retail mobile manipulation such as Spahn et al.~\cite{airlab_paper}, which demonstrates feasibility on higher-end platforms, our solution targets cost-constrained deployment and improves shelf-proximal navigation and manipulation reliability via visual markers plus MPC and improved perception, achieving $98\%$ success over 724 pick-place events with autonomous docking/charging on a commercially available mobile manipulator. Third, we provide a detailed validation in an overnight closed-door operation scenario under standard indoor lighting, matching retailer practice to avoid customer congestion and to open with restocked shelves; and a quantitative outlook, including side-by-side benchmarks against teleoperation and human workers, a joint time-cost performance index, and a hardware/software gap decomposition outlining concrete paths toward commercially viable retail robots.

The paper is organized as follows: Section~\ref{sec:preliminaries} introduces foundational background in BTs, MPC, KF, and vision-based segmentation and classification tools. Section~\ref{sec:system} provides an in-depth breakdown of our hardware setup. Section~\ref{sec:results} describes in detail our algorithm, from planning to manipulation. Finally, Section~\ref{sec:experimental} presents extensive laboratory tests, critically assessing the current potential, and fundamental limitations, of AMRs solutions for widespread implementation in retail store environments, with a major focus on the cost-to-performance benefit compared to human workers.

\begin{figure*}[ht]
    \centering
    %
    \begin{subfigure}[b]{0.19\textwidth}
    \centering
        \includegraphics[width=\textwidth]{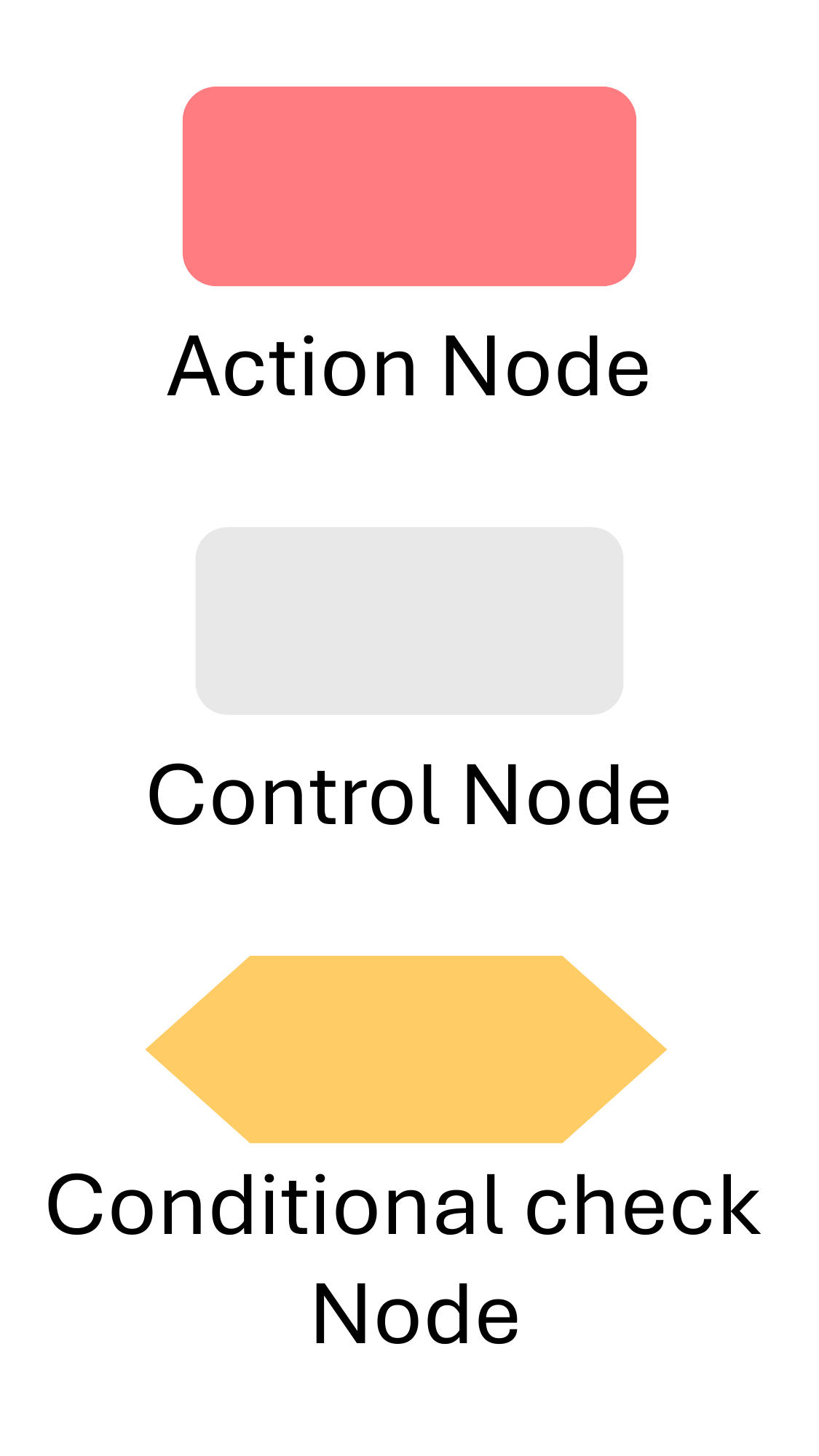}
        \caption{BTs nodes.}
        \label{fig:behaviour_tree_nodes}
    \end{subfigure}
    \hfill
    \begin{subfigure}[b]{0.8\textwidth}
        \centering
        \includegraphics[width=\textwidth]{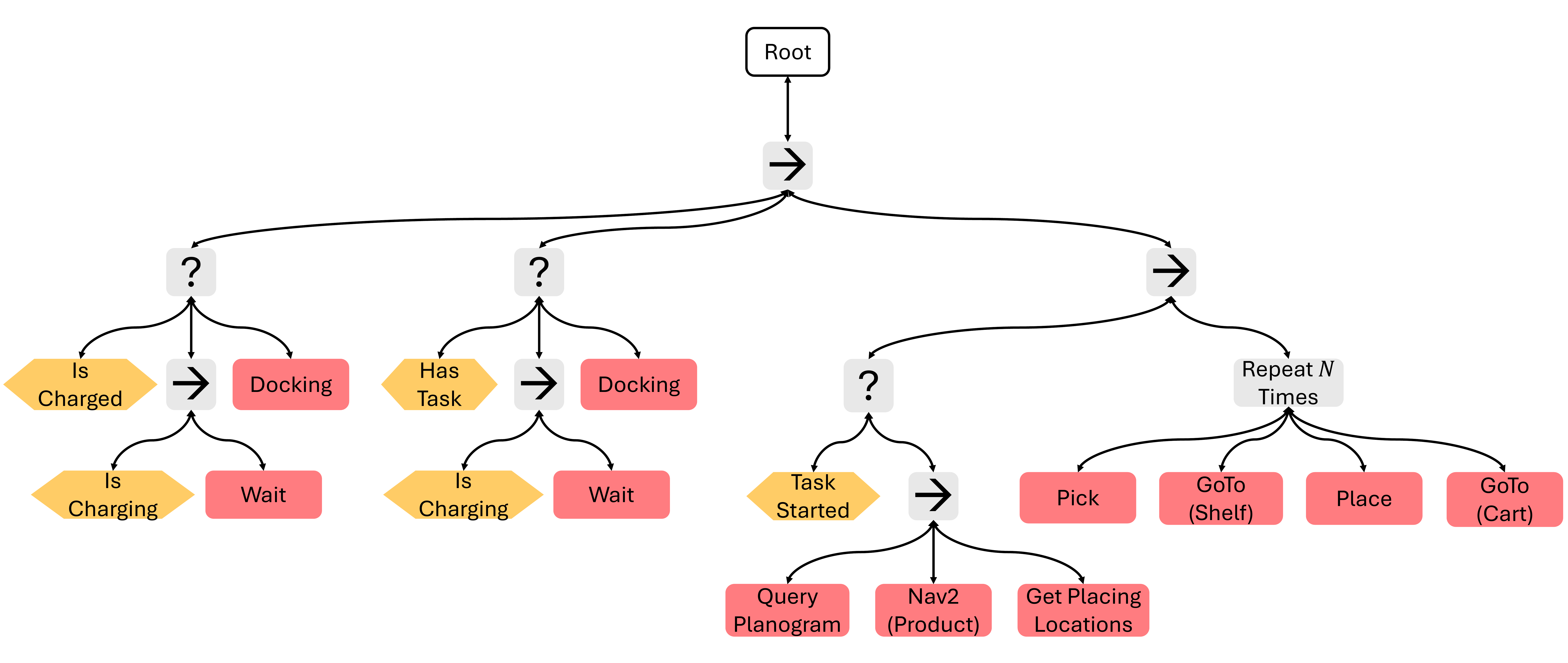}
        \caption{BT developed.}
        \label{fig:behaviour_tree}
    \end{subfigure}
    \caption{Behaviour Trees (BTs).}
\end{figure*}

\section{Preliminaries} \label{sec:preliminaries}
The tools and algorithms below were selected to meet the constraints of retail stocking, i.e., robustness in dynamic spaces, real-time execution on consumer hardware, compatibility with ROS2 and available open-source frameworks, and ease of maintenance. Thus, we focus on the synergetic integration of mature methods widely adopted in robotics.

\subsection{Behavior Trees}
We use Behaviour Trees (BTs) to model complex, high-level decision processes and task planning due to their modular, reactive, and fault-tolerant nature. They are also easy to extend and debug, while being widely used~\cite{nav2}. They consist of three main nodes (Fig.~\ref{fig:behaviour_tree_nodes}): \textbf{action nodes} executing specific actions; \textbf{control nodes} managing the execution flow, including \textit{sequence} nodes (\(\rightarrow\)) that execute child nodes sequentially until one fails, and \textit{fallback} (selector) nodes (?) that do so until one succeeds; and \textbf{conditional check nodes} evaluating conditions to adjust behavior. Combining these nodes enables structured and adaptive decision-making, facilitating the execution of complex tasks in dynamic environments. A detailed algorithmic description of BTs can be found in~\cite{bt_intro}.

\subsection{Model Predictive Control}
We use a Model Predictive Control (MPC) approach as part of our control strategy to solve the "last-meter" navigation problem since it natively enforces motion and safety constraints for coupled robot base-head control while remaining real-time on consumer hardware. Given a control-affine discrete-time system with state vector $\boldsymbol{x}\in \mathcal{X}\subseteq\mathbb{R}^n$, input vector $\boldsymbol{u}\in\mathcal{U}\subseteq\mathbb{R}^m$, and sampling time $T$, the dynamics can be represented as~\eqref{eq:discrete_dynamics}. Consider prediction horizon $N$, input sequence $U=\{\boldsymbol{u}_k,\dots,\boldsymbol{u}_{k+N-1}\}$, and given the generic cost function $c: \mathcal{X}\times\mathcal{U}\rightarrow\mathrm{R}$, the MPC optimization, following~\cite{mpc}, is
\begin{subequations}\label{eq:mpc}
\begin{align}
  \min_{U}&\quad c(\boldsymbol{x},\boldsymbol{u}) \\  
  \text{s.t.}& \quad \boldsymbol{x}_{k+1} = \boldsymbol{x}_{k} + T \cdot f(\boldsymbol{x}_{k}, \boldsymbol{u}_{k}),\label{eq:discrete_dynamics}\\
&\quad\boldsymbol{x}_{k}\in\mathcal{X},\quad \boldsymbol{u}_{k}\in\mathcal{U},\quad k=0,...,N, \\
&\quad g(x,u)\leq0, \quad \boldsymbol{x}_0=\tilde{x}_0,
\end{align}
where $g(x,u)$ represents joint state and input constraints, while $\boldsymbol{x}_0=\tilde{x}_0$ enforces the initial state to be the measured state $\tilde{x}_0$. The solution provides optimal control inputs satisfying navigation requirements and imposed constraints.
\end{subequations}

\subsection{Kalman Filter}
To accurately locate the robot within cluttered and partially observable environments, such as grocery store aisles, reliable pose estimation using visual markers is essential. Consider a discrete-time linear system
\begin{equation}\label{eq:discrete_LTS}
\boldsymbol{x}_{k+1} = A\boldsymbol{x}_k + B\boldsymbol{u}_k + \boldsymbol{w}_k,\quad \boldsymbol{z}_k = H\boldsymbol{x}_k + \boldsymbol{v}_k,
\end{equation}
where $\boldsymbol{z}_k$ is the measurement, and $\boldsymbol{w}_k \sim \mathcal{N}(0,Q)$ and $\boldsymbol{v}_k \sim \mathcal{N}(0,R)$ are zero-mean Gaussian process and measurement noise, respectively. The Kalman Filter (KF) performs a two-step predict-update cycle, which is used to estimate noisy processes, e.g., visual pose estimation, reliably:
\begin{align}
\text{Predict:} \quad & \hat{\boldsymbol{x}}_{k|k-1} = A\hat{\boldsymbol{x}}_{k-1|k-1} + B\boldsymbol{u}_{k-1}, \\
& P_{k|k-1} = AP_{k-1|k-1}A^\top + Q, \\
\text{Update:} \quad & K_k = P_{k|k-1}H^\top(H P_{k|k-1} H^\top + R)^{-1}, \\
& \hat{\boldsymbol{x}}_{k|k} = \hat{\boldsymbol{x}}_{k|k-1} + K_k(\boldsymbol{z}_k - H\hat{\boldsymbol{x}}_{k|k-1}), \\
& P_{k|k} = (I - K_k H)P_{k|k-1}.
\end{align}
In our use case, a KF is suitable since, when combined with ArUco markers observations, they ensure accurate localization even in close-range settings where lidar-based SLAM fails due to its minimum range limitation. For further details on the KF, the reader is encouraged to refer to \cite{Kalman_1960}.

\subsection{Visual Models}
\subsubsection{YOLO} (You Only Look Once)~\cite{yolo} is a state-of-the-art Convolutional Neural Network (CNN) based object detection framework that performs bounding box and class predictions in a single forward pass. We used YOLO11n-cls \cite{YOLO11} for its strong speed/accuracy trade-off and ease of fine-tuning with a modest amount of data required.
\subsubsection{SAM2} (Segment Anything Model 2)~\cite{sam2} is a transformer-based architecture for promptable visual segmentation, designed to extend image segmentation capabilities to the video domain.  We used it due to its combined masking and video tracking capabilities and its ability to run in real-time. We use two instances of SAM2 during picking: Alg.~\ref{alg:pick}, \SAMSegment for scene segmentation within an image (Fig.~\ref{fig:pick_sam2}), and \SAMtrack for object tracking between consecutive frames (Fig.~\ref{fig:pick_track}).


\section{System Overview} \label{sec:system}
This section describes the components of our system and the supermarket setting used for experimental validation.

\subsection{Hardware Overview}
The system comprises a mobile robot and an external computer for more complex perception-related computations, both running ROS2~\cite{ros2} and connected to the same Wi-Fi.

\begin{figure}[h]
    \centering
    %
    \begin{subfigure}[b]{0.25\linewidth}
        \centering
        \includegraphics[width=\textwidth]{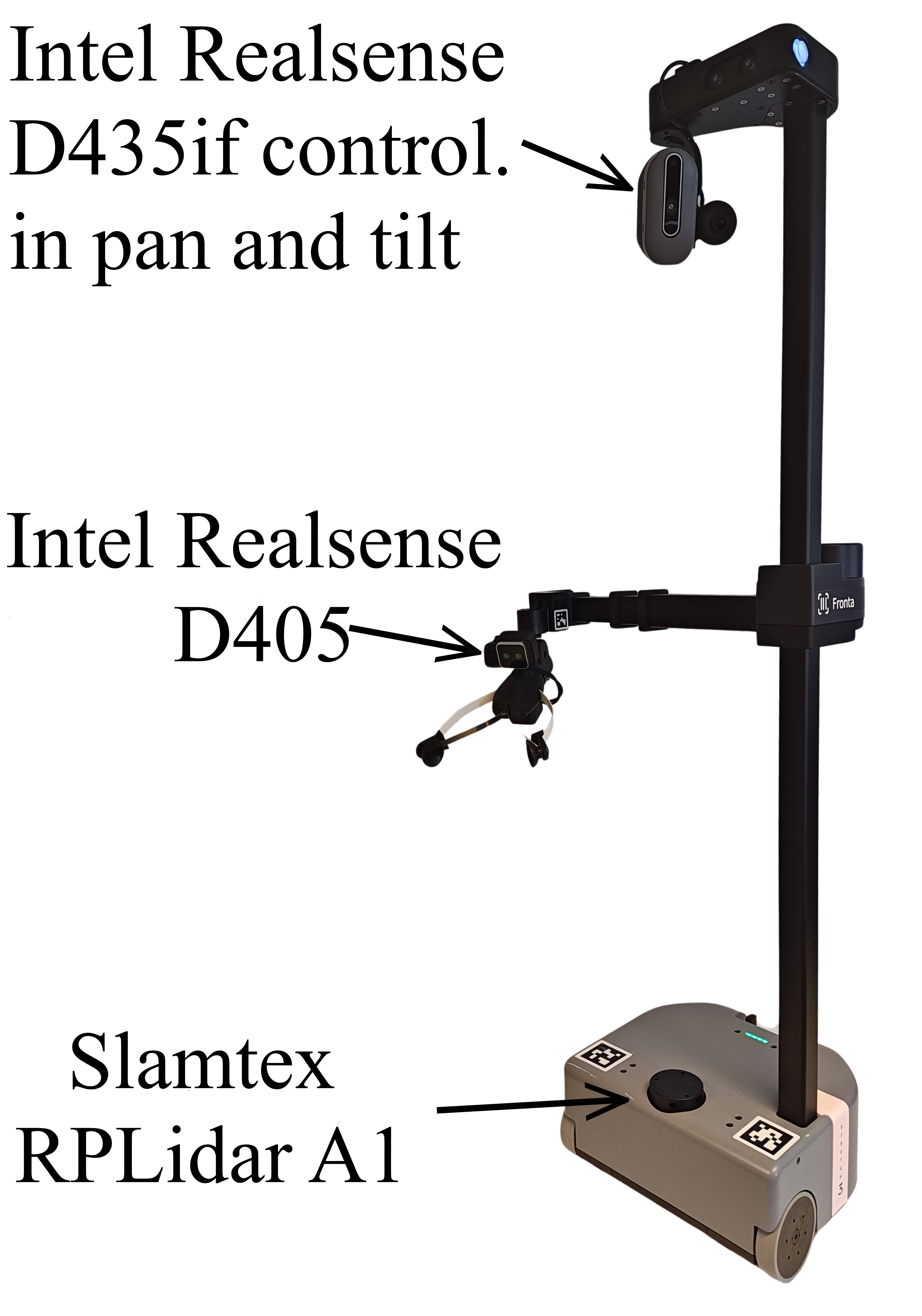}
        \caption{Stretch 3.}
        \label{fig:robot}
    \end{subfigure}
    \hfill
    \begin{subfigure}[b]{0.63\linewidth}
    \centering
    \includegraphics[width=\textwidth]{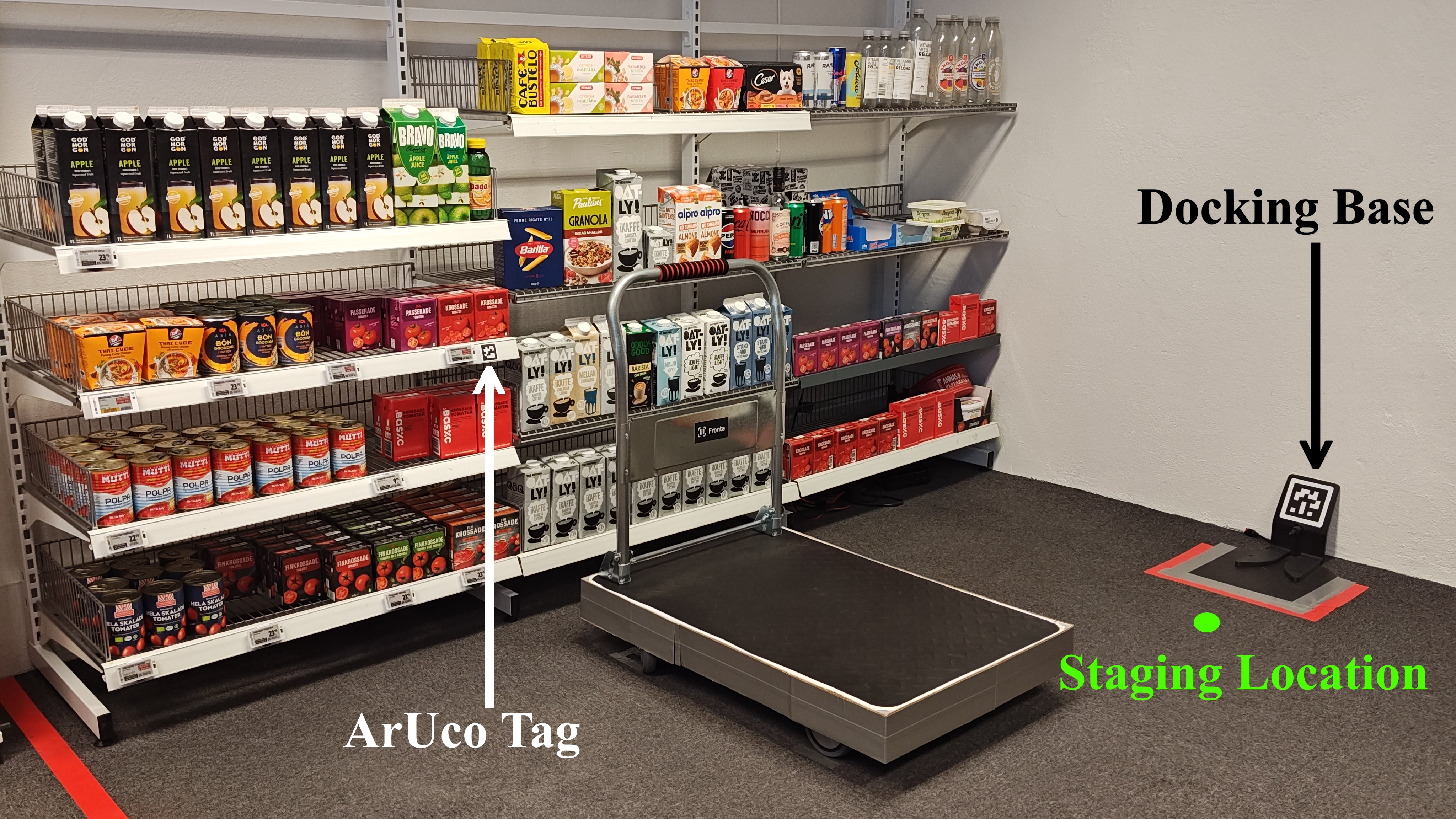}
    \caption{Mock supermarket setup.}
    \label{fig:supermarket}
    \end{subfigure}
    \caption{Hardware overview.}
\end{figure}

We use the  Hello Robot Stretch 3 (Fig.~\ref{fig:robot}) which is a mobile manipulator with a differential drive base modeled as a unicyle with state $\boldsymbol{x}_b=\left[x_b, y_b,\theta\right]^T$ representing its position and orientation, control input $\boldsymbol{u}_b=\left[v,\omega\right]^T$ representing its linear and angular velocity, and dynamics
\begin{equation}\label{eq:unicycle}
   {\boldsymbol{x}}_{b,k+1}= \boldsymbol{x}_{b,k}+T\cdot\left[v_k \cos{\theta_k}, v_k \sin{\theta_k}, \omega_k\right]^T,
\end{equation} 
 and a head camera modeled as an integrator with state $\boldsymbol{x}_h=\left[p, t\right]^T$ representing the pan and tilt respectively, input $\boldsymbol{u}_h=\left[v_p, v_t\right]^T$ representing the associated velocities, and dynamics \begin{equation}\label{eq:head_dyn}
      \boldsymbol{x}_{h,k+1}=\boldsymbol{x}_{h,k}+T\cdot\left[v_{p,k},  v_{t,k}\right]^T.
  \end{equation} 
  
Stretch 3 also has a vertical lift mechanism, a telescopic arm, and a wrist with 3 degrees of freedom (DoF), allowing object manipulation at various distances and heights. It is powered by an Intel NUC12WSKi5 to run planning and navigation nodes. For technical specifications, refer to the datasheet provided by the manufacturer \cite{stretch}.

The workstation is powered by an Intel Core i9-14900 CPU and an NVIDIA RTX 4090 GPU, enabling real-time execution of computationally intensive perception algorithms. 
\subsection{Supermarket Setting}
We performed experiments within a mock supermarket environment (Fig.~\ref{fig:supermarket}). The products to be stocked are placed on a mobile cart located near the target shelves. While the system navigates between them, this work focuses on the manipulation pipeline, assuming the cart's initial placement is within the local navigation frame. To guarantee a reliable pose estimation without affecting realistic store operations, we placed small ArUco tags in the store shelf label holders. These visual markers, read by the head camera, provide a stable reference frame used within the KF, enabling the robot to accurately locate its pose $\boldsymbol{x}_b$~\cite{aruco} in the world frame thanks to the prior knowledge of the ArUco markers' pose in this world reference frame.
We use ArUco tags because they provide absolute 6-DoF anchors that are robust to lighting, occlusion, and run in real-time on CPU. Moreover, impact on shelves is negligible: $35\si{mm}$ paper tags slide into standard shelf-label rails ($\approx39\si{mm}$), sit flush behind the clear plastic with prices still visible, are installed or removed in seconds, and can be registered within the system by a brief scan.


\section{Main Results}\label{sec:results}
Our main contribution stands in a deployable ROS2 pipeline that tightly integrates standard methods with task-specific modules: a ROS2 BT with Message Queuing Telemetry Transport (MQTT)~\cite{mqtt} integration for task generation and visualization; a Spatio-Temporal Voxel Layer (STVL)~\cite{stvl} layer and an ArUco-based visual docking added to the standard Nav2; and a visual shelf-navigation stack combining a 14-state KF with a two-step MPC. For perception, SAM2 masks and tracking are paired with per-SKU YOLO fine-tunes and geometry-aware mask selection; for manipulation, we use centroid-based velocity control and planogram-aware Inverse Kinematics (IK) placement with contact detection. Many of these components are open-sourced as ROS2 packages~\cite{our_docking, our_nav2, ros_bridge, our_simulator}.

\subsection{Planning}\label{sec:planning}

\begin{figure}[h]
    \centering    \includegraphics[width=0.91\linewidth]{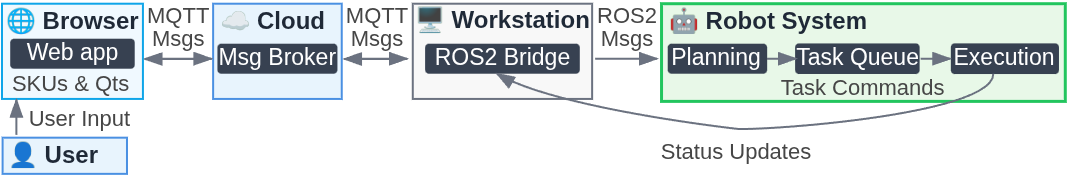}
    \caption{Planning and communication diagram.}
    \label{fig:planning}
\end{figure}

Our planning architecture operates through a two-step concurrent process designed to dynamically manage the robot's behavior (Fig.~\ref{fig:planning}). At the higher level, tasks are assigned to the robot via our web application: it allows users to specify products and quantities to restock, generating corresponding tasks transmitted via MQTT~\cite{mqtt}. Tasks reach the robot's planning node through our open source ROS2-MQTT bridge~\cite{ros_bridge}, maintaining a real-time adjustable task queue.
The BT in Fig.~\ref{fig:behaviour_tree} handles the lower-level recurrent planning. It is composed of three subtrees: the leftmost and middle ones for docking when charging is needed and when the robot is idle, respectively; and the right one to complete the assigned tasks from the MQTT task queue. Most BT actions are explained in the following sections. The \textit{Wait} action indicates robot idle periods, while conditional check nodes are implemented via boolean variables, always accessible to the planning node.
\subsection{Navigation}\label{sec:nav2}
\begin{figure}[b]
    \centering
    \includegraphics[width=\linewidth]{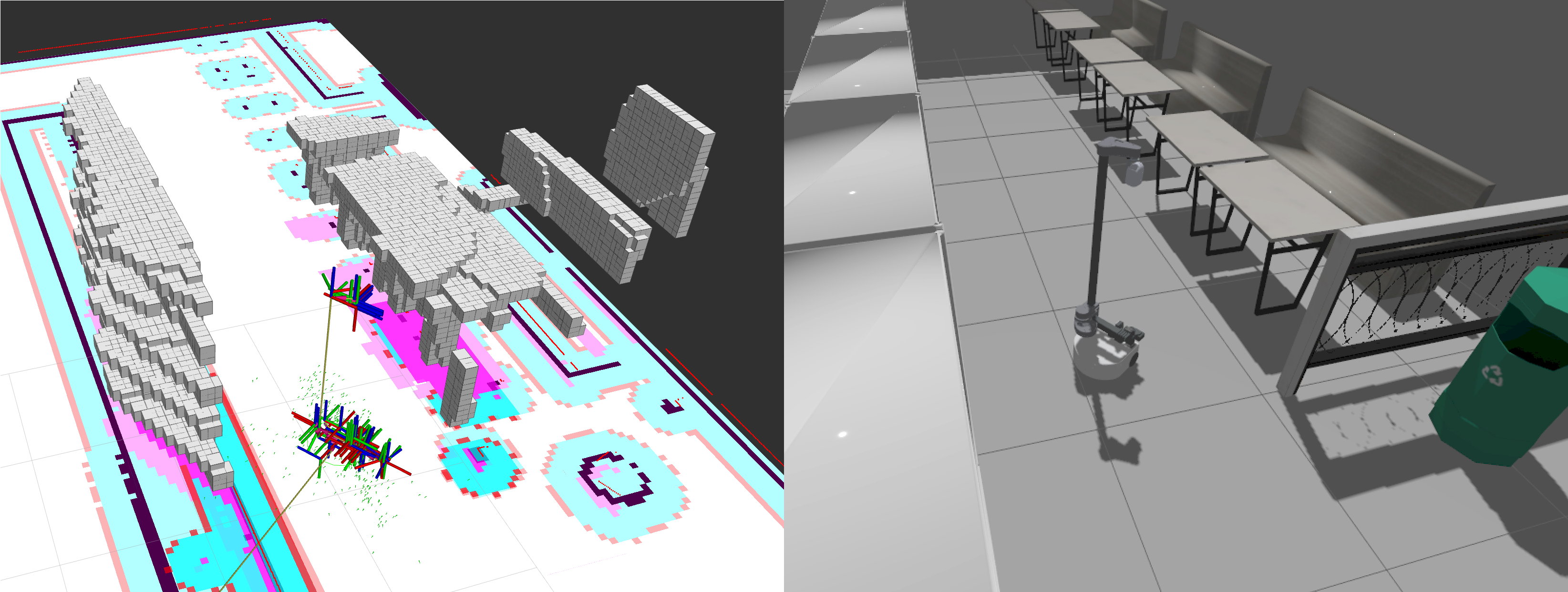}
    \caption{Nav2 interface (left) and the simulator (right).}
    \label{fig:nav2}
\end{figure}

Our open-source navigation module \cite{our_nav2} uses the open-source ROS2 Nav2 package~\cite{nav2}, which provides a modular framework for autonomous robot navigation that integrates various planners and controllers. We use a Model Predictive Path Integral (MPPI) controller~\cite{mppi}, which ensures smooth, real-time trajectories. Additionally, we introduced a STVL~\cite{stvl} to the navigation stack to improve obstacle perception. This layer, which can be seen in action in Fig.~\ref{fig:nav2} left, uses stereo-depth data from the head camera to detect obstacles imperceptible to lidar, e.g., elevated objects, significantly improving navigation safety.
\begin{remark}
Prior to operation, the environment is mapped using the ROS2 SLAM Toolbox~\cite{slam}, enabling autonomous navigation by specifying target locations within the map.
\end{remark}
\subsubsection{Docking}\label{sec:docking} 
This action, available at \cite{our_docking}, initiates an automatic docking sequence. Initially, the robot autonomously navigates to a staging location in front of the charging base using Nav2 (Fig.~\ref{fig:supermarket}). Next, the robot rotates $180^\circ$, aligning its base with the charging dock and positioning the camera to face backward. Lastly, a visual servo controller~\cite{visual_servo}, using real-time ArUco tag detection, computes target coordinates as $(x_{des}, y_{des}) = Transform(x_{aruco}, y_{aruco})$, converting in real-time the ArUco position from the head camera frame to the desired robot base position in the world frame. The controller uses a proportional control law to generate base velocities: $v_b = K_{x}(x_{curr} - x_{des})$ and $\omega_b = K_{y}(y_{curr} - y_{des})$ with  $K_x,K_y\in\mathbb{R}^+$. If visual servoing fails, e.g., due to loss of the ArUco marker, Nav2 re-engages to position the robot to the staging location. This reliably allows the robot to dock, recharge, and autonomously resume or start new tasks.

\begin{figure*}[ht]
    \centering
    %
    \begin{subfigure}[b]{0.195\textwidth}
        \centering
        \includegraphics[width=\textwidth]{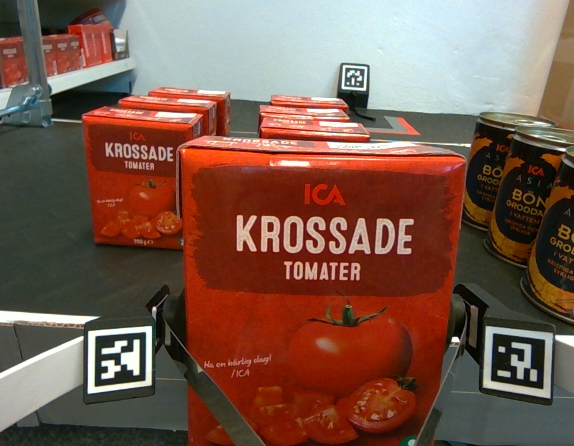}
        \caption{Picking task (Fig.~\ref{fig:pick_task}).}
        \label{fig:picking}
    \end{subfigure}
    \hfill
    \begin{subfigure}[b]{0.195\textwidth}
        \centering
        \includegraphics[width=\textwidth]{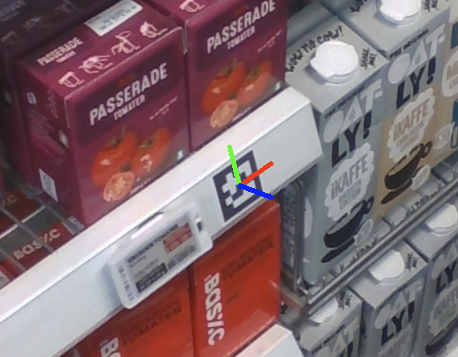}
        \caption{Navigate to shelf.}
        \label{fig:micro_navigation1}
    \end{subfigure}
    \hfill
    \begin{subfigure}[b]{0.195\textwidth}
        \centering
        \includegraphics[width=\textwidth]{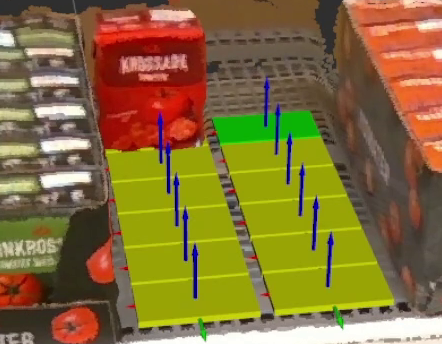}
        \caption{Detect place location.}
        \label{fig:place_perception}
    \end{subfigure}
    \hfill
    \begin{subfigure}[b]{0.195\textwidth}
        \centering
        \includegraphics[width=\textwidth]{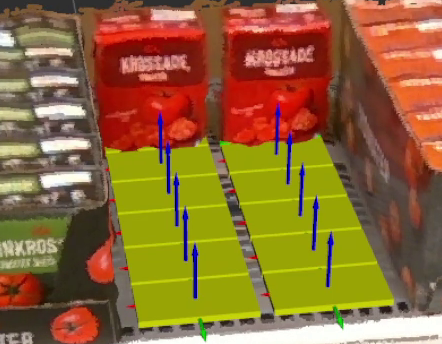}
        \caption{Place using IK.}
        \label{fig:place_IK}
    \end{subfigure}
    \hfill
    \begin{subfigure}[b]{0.195\textwidth}
        \centering
        \includegraphics[width=\textwidth]{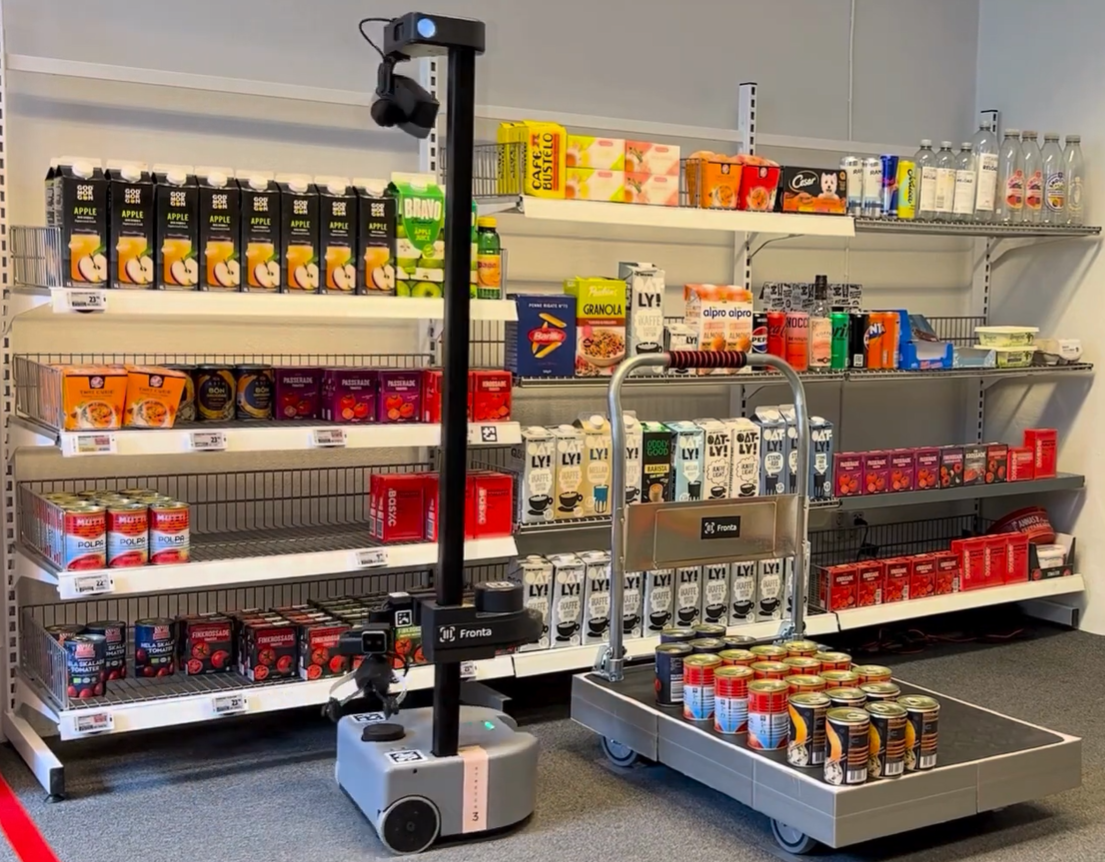}
        \caption{Navigate to cart.}
        \label{fig:micro_navigation_2}
    \end{subfigure}

    \caption{Overview of the pick and place task.}
    \label{fig:pick_place_task}
\end{figure*}

\begin{figure*}[ht]
    \centering
    %
    \begin{subfigure}[b]{0.195\textwidth}
        \centering
        \includegraphics[width=\textwidth]{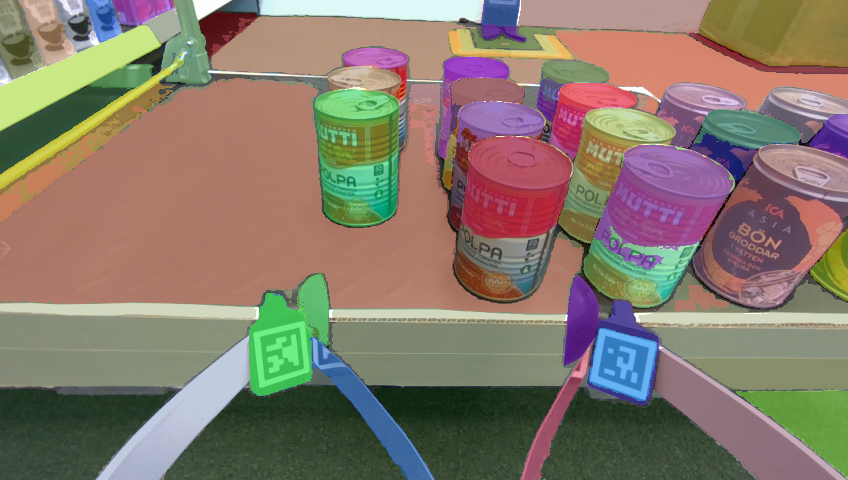}
        \caption{Segment RGB image.}
        \label{fig:pick_sam2}
    \end{subfigure}
    \hfill
    \begin{subfigure}[b]{0.195\textwidth}
        \centering
        \includegraphics[width=\textwidth]{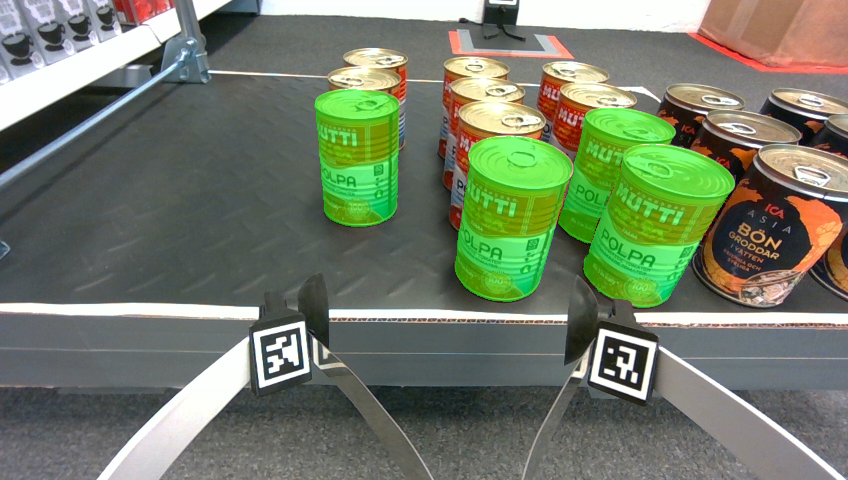}
        \caption{Detect correct products.}
        \label{fig:pick_yolo}
    \end{subfigure}
    \hfill
    \begin{subfigure}[b]{0.195\textwidth}
        \centering        \includegraphics[width=\textwidth]{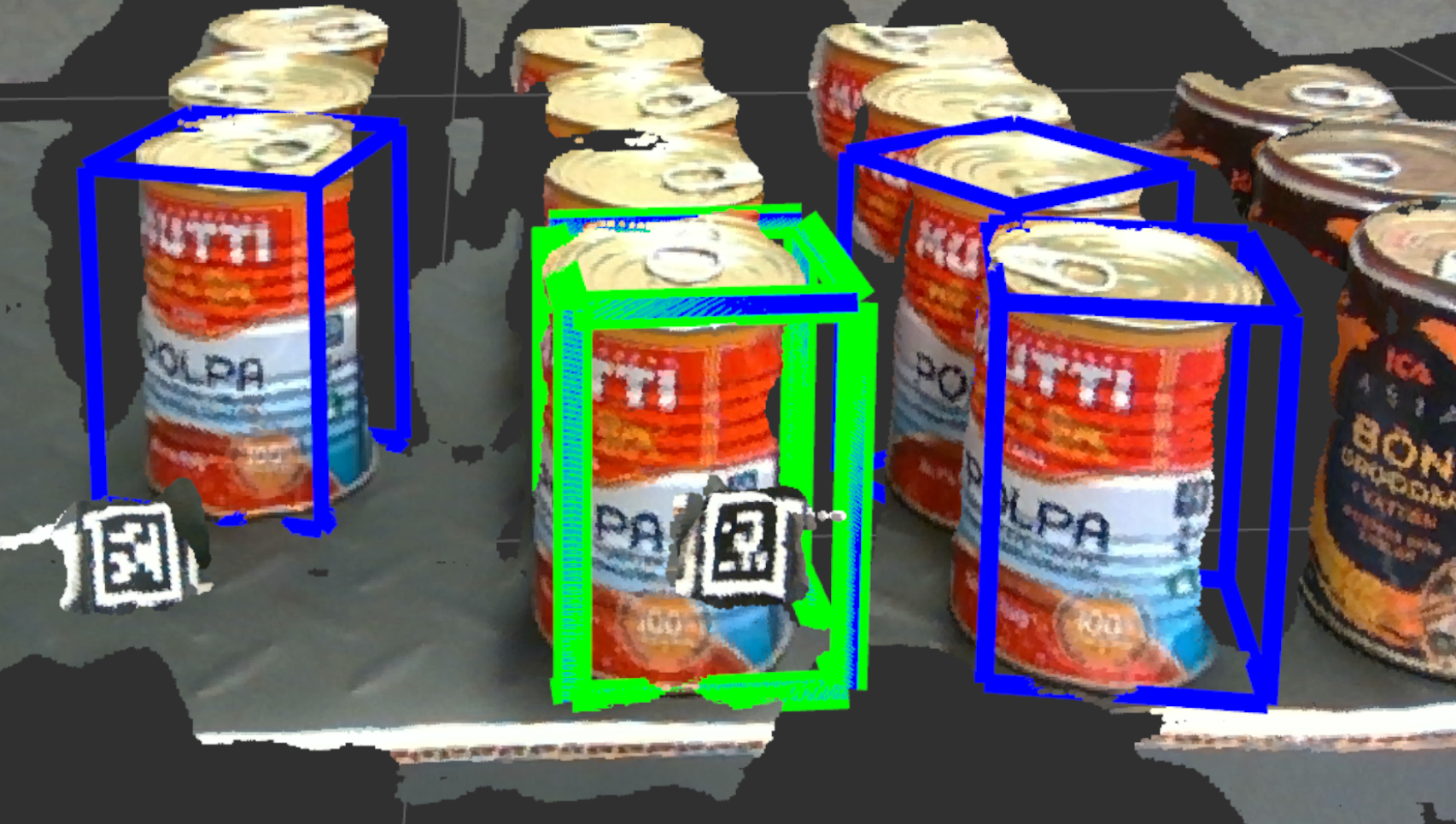}
        \caption{Bounding box select.}
        \label{fig:pick_bounding_box}
    \end{subfigure}
    \hfill
    \begin{subfigure}[b]{0.195\textwidth}
        \centering
        \includegraphics[width=\textwidth]{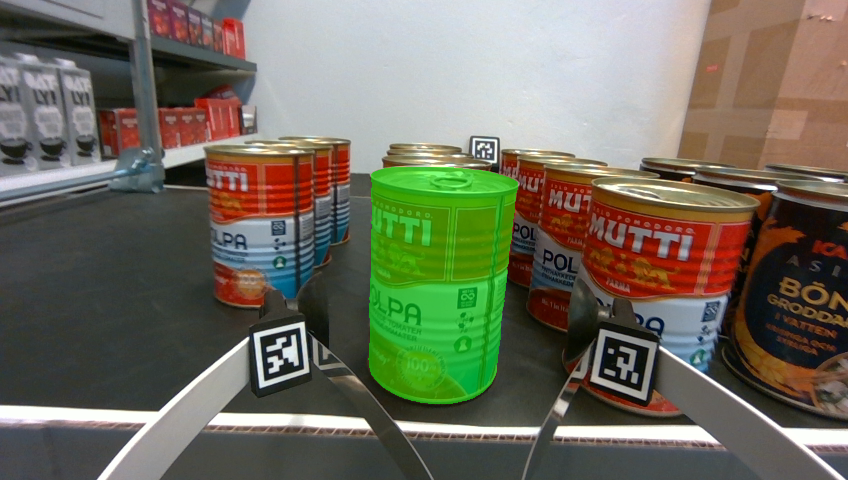}
        \caption{Track selected product.}
        \label{fig:pick_track}
    \end{subfigure}
    \hfill
    \begin{subfigure}[b]{0.195\textwidth}
        \centering
        \includegraphics[width=\textwidth]{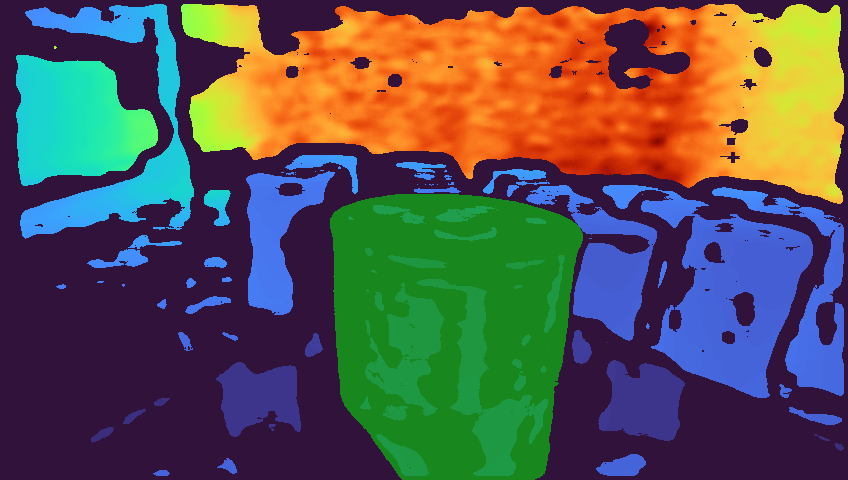}
        \caption{Pick using depth image.}
        \label{fig:pick_pointcloud}
    \end{subfigure}

    \caption{Overview of the perception pipeline used for picking.}
    \label{fig:pick_task}
\end{figure*}
\subsection{Visual shelf-navigation}\label{sec:micro_navigation}

\begin{problem}["last-meter" Navigation]
The lidar's minimum detection range ($20\si{cm}$) prevents accurate sensing of objects closer than this limit, such as shelves ($\approx10\si{cm}$ away), requiring an alternative approach for shelf navigation.
\end{problem}
To solve this problem, we combine visual markers, a Kalman Filter, and a two step Model Predictive Controller. 

\subsubsection{Kalman Filter}\label{sec:visual_kalman}
To improve the accuracy and robustness of the ArUco-based localization, we implemented a 14-dimensional KF tailored to the 6-DoF pose estimation of the marker detected through the head camera. The filter state vector is defined as $\boldsymbol{x} = [\boldsymbol{p}, \boldsymbol{q}, \dot{\boldsymbol{p}}, \dot{\boldsymbol{q}}]^\top \in \mathbb{R}^{14},$ where $\boldsymbol{p} = [x, y, z]^\top$ and $\boldsymbol{q} = [q_x, q_y, q_z, q_w]^\top$ represent the position and orientation quaternion, respectively, and $\dot{\boldsymbol{p}}, \dot{\boldsymbol{q}}$ denote their velocities. The filter employs a constant velocity model and discrete-time dynamics~\eqref{eq:discrete_LTS} with
\begin{equation}
     A=\begin{bmatrix}
\mathbf{I}_7 & \Delta t \cdot \mathbf{I}_7 \\
\mathbf{0}_{7 \times 7} & \mathbf{I}_7
\end{bmatrix},
\end{equation}
where $\Delta t $ is the sampling time and $\Delta t \cdot \mathbf{I}_7$ is used to couple position/orientation with their respective velocities; $B=0_{14\times m}$, and $H \in \mathbb{R}^{7 \times 14}$ extracts the observable components $\boldsymbol{z}_k = [\boldsymbol{p}, \boldsymbol{q}]^\top$ from the state. In this work quaternion normalization is enforced post-update.

\subsubsection{Two Step MPC}
Now that we have an accurate way to estimate the robot's pose we use a two step MPC to move it to the desired location. For the base MPC we consider the dynamics \eqref{eq:unicycle}, and the following cost function 
\begin{equation}
    \sum\limits^{N-1}_{i=0} (\boldsymbol{x}_{b, k+i}-\boldsymbol{x}_d)^TQ(\boldsymbol{x}_{b, k+i}-\boldsymbol{x}_d)+ {\boldsymbol{u}_{b,k+i}}^TR {\boldsymbol{u}_{b, k+i}},
\end{equation}
where $\boldsymbol{x}_d$ is the desired pose and $Q\in\mathbb{R}^{3\times3}$, $R\in\mathbb{R}^{2\times2}$ are used for tuning the controller. We also limit the robot workspace $\mathcal{X}_b$ and the input space $\mathcal{U}_b$.
Let $X_b$ be the sequence of states outputted for the base MPC, which will be used as a parameter to calculate the camera pose for the second step.

The next step is the head MPC with the dynamics~\eqref{eq:head_dyn} and an input smoothness cost function with weight $K\in\mathbb{R}^{2\times2},$
\begin{equation}
    \sum\limits^{N}_{k=1} (\boldsymbol{u}_{h, k}-\boldsymbol{u}_{h, k-1})^TK(\boldsymbol{u}_{h, k}-\boldsymbol{u}_{h, k-1}).
\end{equation}
We limit both the workspace $\mathcal{X}_c$ and the input space $\mathcal{U}_c$ to satisfy joints' limits and add a safety constraint to maintain the ArUco marker within the camera field of view (see~\cite[Eq.~(6)]{visual_servo_victor} and detailed description therein).

\subsection{YOLO Fine-tuning}\label{subsec:yolo}
To categorize and select the products to pick in some instances, we rely on fine-tuning a pre-trained YOLO11n-cls weights from ImageNet, fully updating all layers to specialize in grocery product identification with a similar procedure to~\cite{yolo_finetuning}. 
From this baseline, we trained a grocery-specific version, \yologrocery, to guarantee fast and accurate detection (in Alg.~\ref{alg:pick}). The training involved 200 epochs with $96\times96$ input resolution using a dataset, which also included slightly damaged goods, created during robot execution, batch size of 640, and Adam optimizer (learning rate: 0.005 to 0.0001, with warmup scheduling). To mitigate overfitting, we incorporate label smoothing (0.1) and random erasing. Lastly, we trained multiple product-specific fine-tuned models, \selectprodyolo, trained on smaller datasets with a procedure similar to the one just explained. These models are used to detect the products the robot will pick (Fig.~\ref{fig:pick_yolo}).

\subsection{Manipulation}

In this section, we will go over the manipulation elements of our pipeline, which include the picking and placing actions and their relative perception pipelines and controllers.

\subsubsection{\textbf{Pick}}\label{sec:pick}
The picking pipeline (Fig.~\ref{fig:pick_task}, Alg.~\ref{alg:pick}) requires the target product, identified via a Global Trade Item Number (GTIN)~\cite{gtin}, and a boolean flag $has\_prod\_YOLO$ indicating the availability of a product-specific YOLO model (Sec.~\ref{subsec:yolo}). These models may not be available for all products due to the effort needed to create datasets. Initially, the robot takes a snapshot, segments the scene using SAM2 (Fig.~\ref{fig:pick_sam2}), and filters segments with the grocery YOLO model (Fig.~\ref{fig:pick_yolo}). If $has\_prod\_YOLO$ is true, a product-specific YOLO identifies the target product segments; otherwise, GPT-4o~\cite{gpt4} is used with a reference image and a carefully designed prompt. If detection fails, the robot randomly repositions and retries up to three times before returning $False$. From valid segments, point clouds are generated, and known product dimensions from the database guide an optimization that computes accurate bounding boxes (Fig.~\ref{fig:pick_bounding_box}), further filtered using visibility and graspability criteria. A SAM2-based tracker initialized on the selected product's mask aligns the robot base with the object's centroid (Fig.~\ref{fig:pick_track}), while real-time point cloud data informs a velocity-based arm controller (Fig.~\ref{fig:pick_pointcloud}). Given object centroid $[x_o, y_o]$, target grasp position $[x_t, y_t]$ within the image and the gains $P_b,P_l,P_a\in\mathbb{R}^+$, the velocities are computed as $v_b = P_b(x_t - x_o)$ for the base and $v_l = P_l(y_t - y_o)$ for vertical lift. Finally, object's distance $z_o$ from the camera, derived as the $99$-th percentile depth of masked points, determines arm velocity as $v_a = P_a(z_o - l_g)$, with $l_g$ being the camera-to-gripper distance, know a priori by measuring it on the robot.

\begin{algorithm2e}[ht]
\DontPrintSemicolon
\SetKwFunction{selectprodllm}{GPT4o}
\SetKwFunction{getprodinfo}{Get\_Prod\_Info}
\SetKwFunction{adjustbase}{Adjust\_Base}
\SetKwFunction{snapshot}{Take\_Snapshot}
\SetKwFunction{pointcloud}{Get\_Pointclouds}
\SetKwFunction{boundingbox}{Get\_Bounding\_Boxes}
\SetKwFunction{filterbbs}{Filter\_Bounding\_Boxes}
\SetKwFunction{adjustbasearm}{Adjust\_Base\_Arm}
\SetKwFunction{rand}{Rand}
\SetKwFunction{grasp}{Close\_Gripper}
\SetKw{Continue}{continue}
\SetKwFunction{planogram}{Query\_Planogram}
\SetKwFunction{place}{Get\_Place\_Poses}
\SetKwFunction{IK}{IK\_Control}
\SetKwFunction{aruco}{ArUco\_Est}
\SetKwFunction{arm}{Lower\_Arm}
\SetKwFunction{leave}{Open\_Gripper}
\SetKwFunction{home}{Home\_Arm}
\SetKwFunction{base}{Move\_Base}

\SetKwInOut{Input}{Input}
\SetKwInOut{Output}{Output}
\ResetInOut{Output}

\Input{$GTIN$, $has\_prod\_YOLO$}
\Output{$success$}
\BlankLine
$RGBD\_img=$\snapshot{}\;
$prod\_info=$\getprodinfo{$GTIN$}\;

\For{$i\in[0,2]$}{
    $color=RGBD\_img.color\_img$\;
    $depth=RGBD\_img.depth\_img$\;
    $segments=$\SAMSegment{$color$}\;
    $segments=$\yologrocery{$segments$}\;
    \If{$has\_prod\_YOLO$}{
        $segments=$\selectprodyolo{}
    }
    \Else{
        $segments=$\selectprodllm{$prod\_info.img$}
    }
    \If{$segments=\emptyset$}{
        $dist_{meters}=$\rand{$-1,1$}\;
        \adjustbase{$dist_{meters}$}\;
        $RGBD\_img=$\snapshot{}\;
        \Continue
    }
    $pcs=$\pointcloud{$depth$, $segments$}\;
    $bbs=$\boundingbox{$pcs$}\;
    $fin\_seg=$\filterbbs{$bbs$}\;    $mask\_track=$\SAMtrack{$fin\_seg$}\;
    \adjustbasearm{$mask\_track$}\;
    \grasp{}\;
    \Return{True}
}
\Return{False}

\caption{Pick}\label{alg:pick}
\end{algorithm2e}
\subsubsection{\textbf{Place}}\label{sec:place}

It (Alg.~\ref{alg:place}) begins by querying the store's planogram (\planogram) to locate the designated shelf position for the given GTIN. Using an ArUco tag as reference (\aruco), the robot moves to the approximate shelf area. A snapshot of the shelf informs an optimization step (\place) to identify optimal, sequential placement positions. By calculating these poses at the forward-most available coordinates based on product geometry, we place items close to the shelf edge during the stocking process (Fig.~\ref{fig:place_perception}). Next, the robot realigns itself (\base), and an IK controller, similar to~\cite{ik_stretch}, positions the gripper safely above the target location. The arm lowers the product until effort sensors signal shelf contact, triggering a gentle product release and return to the home position. Though the final approach is open-loop, placement accuracy and robustness are ensured by the ArUco-based pose estimation (Sec.~\ref{sec:visual_kalman}, Fig.~\ref{fig:place_IK}).

\begin{algorithm2e}[h]
\DontPrintSemicolon
\SetKw{Not}{Not}
\SetKwInOut{Input}{Input}

\Input{$GTIN$}
\BlankLine
$shelf\_loc=$\planogram{$GTIN$}\;
\base{$shelf\_loc$, \aruco{}}\;

$RGBD\_img=$\snapshot{}\;
$prod\_info=$\getprodinfo{$GTIN$}\;
$place\_poses=$\place{$RGBD\_img$, $prod\_info$}\;
\base{$place\_poses\left[0\right]$, \aruco{}}\;
\IK{$place\_poses\left[0\right]$, \aruco{}}\;
\While{\Not $contact\_detected$}{\arm{}}
\leave{}, \home{}\;
\caption{Place}\label{alg:place}
\end{algorithm2e}

\section{Experimental Results}\label{sec:experimental}
 We target overnight stocking with doors closed and store lighting on, which aligns with common retail practice to avoid crowding customers and to have shelves ready at opening. Under this regime, the environment is consistent across stores, i.e., uniform illumination, predictable layouts, and no bystanders, so dynamic agents are limited to other robots or carts, which can be tracked and avoided in the navigation described in section~\ref{sec:nav2}. Our mock supermarket reproduces this context, allowing us to conduct laboratory experiments to validate our pipeline (video~\cite{video}) and compare its benefits and shortcomings to human workers and teleoperation under realistic conditions.

\subsection{Simulator}
To enable safe and efficient testing, we extended the simulator framework of~\cite{simulator} and developed an open-source, customized Gazebo environment specifically for the Stretch 3, available at~\cite{our_simulator} and shown in Fig.~\ref{fig:nav2} right. The simulator interface and physical robot model were adapted to closely match the actual robot, allowing streamlined development.

\subsection{Experimental Framework}
We validated our approach with tests conducted in a mock supermarket environment set in our laboratory (Fig.~\ref{fig:supermarket}), and then evaluating the success rate and stocking time per object. Due to gripper size ($2.5\si{cm}$ finger width) and payload limitations, we focused on cans and small boxes (side grasps), which represent a large part of the products sold~\cite{canned_food}. Moreover, placement required $3\si{cm}$ spacing between  items, exceeding the common practice of approximately $3\si{mm}$ and limiting dense packing; however, cylindrical items allowed tighter packing ($\approx5\si{mm}$) by adjusting IK approach angles.

Control parameters for visual servoing (Sec.~\ref{sec:docking}) were set at $K_{p,x} = 0.75$, $K_{p,y} = -5.0$, with velocity limits $v_{x,\text{max}}=0.25\si{m/s}$ and $\omega_{z,\text{max}}=2.5\si{rad/s}$ to yield a well-damped response while respecting the robot's velocity limits. For the KF (Sec.~\ref{sec:visual_kalman}), implemented at $30\si{Hz}$ ($\Delta t=1/30$), the noise covariances were estimated from a short static log of ArUco pose jitter to be $\sigma_x=\sigma_z=0.5$; while the states were initialized at first marker detection. The MPC used a sampling time $T=0.05\si{s}$; the base MPC included constraints, at each time step, to prevent collisions with the cart ($x_{b,k+i}\leq 0$), shelf ($y_{b,k+i}\leq 0$), and maintained orientation limits ($\theta_{d}-\epsilon \leq \theta_{k+i}\leq \theta_{d}+\epsilon$), minimizing camera motion blur due to rolling shutter. Lastly, the  gains from Sec.~\ref{sec:pick} were tuned through a trial and error process on the robot to $P_b=0.0005$, $P_l=0.003$, and $P_a=0.9$.

\subsection{Reliability}

Our reliability evaluation focused on two aspects: the success rates of individual stocking operations and overall tasks, and the accuracy of the perception pipeline in categorization.

\begin{figure}[ht]
    \centering
    %
    \begin{subfigure}[b]{1.0\linewidth}
        \centering
        \includegraphics[width=\linewidth]{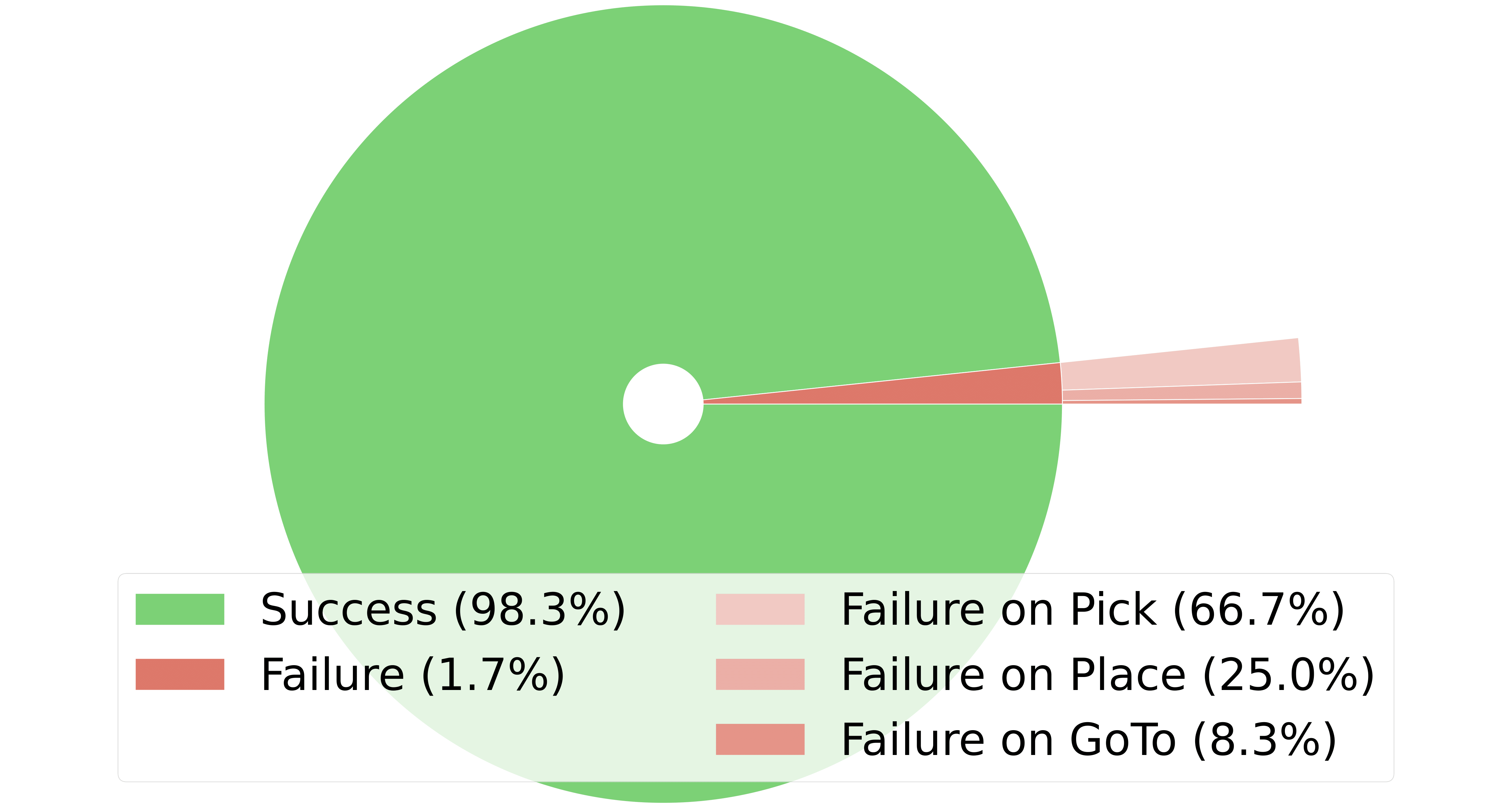}
        \caption{Object level.}
        \label{fig:object_success}
    \end{subfigure}
        \hfill
    \begin{subfigure}[b]{1.0\linewidth}
        \centering
        \includegraphics[width=\linewidth]{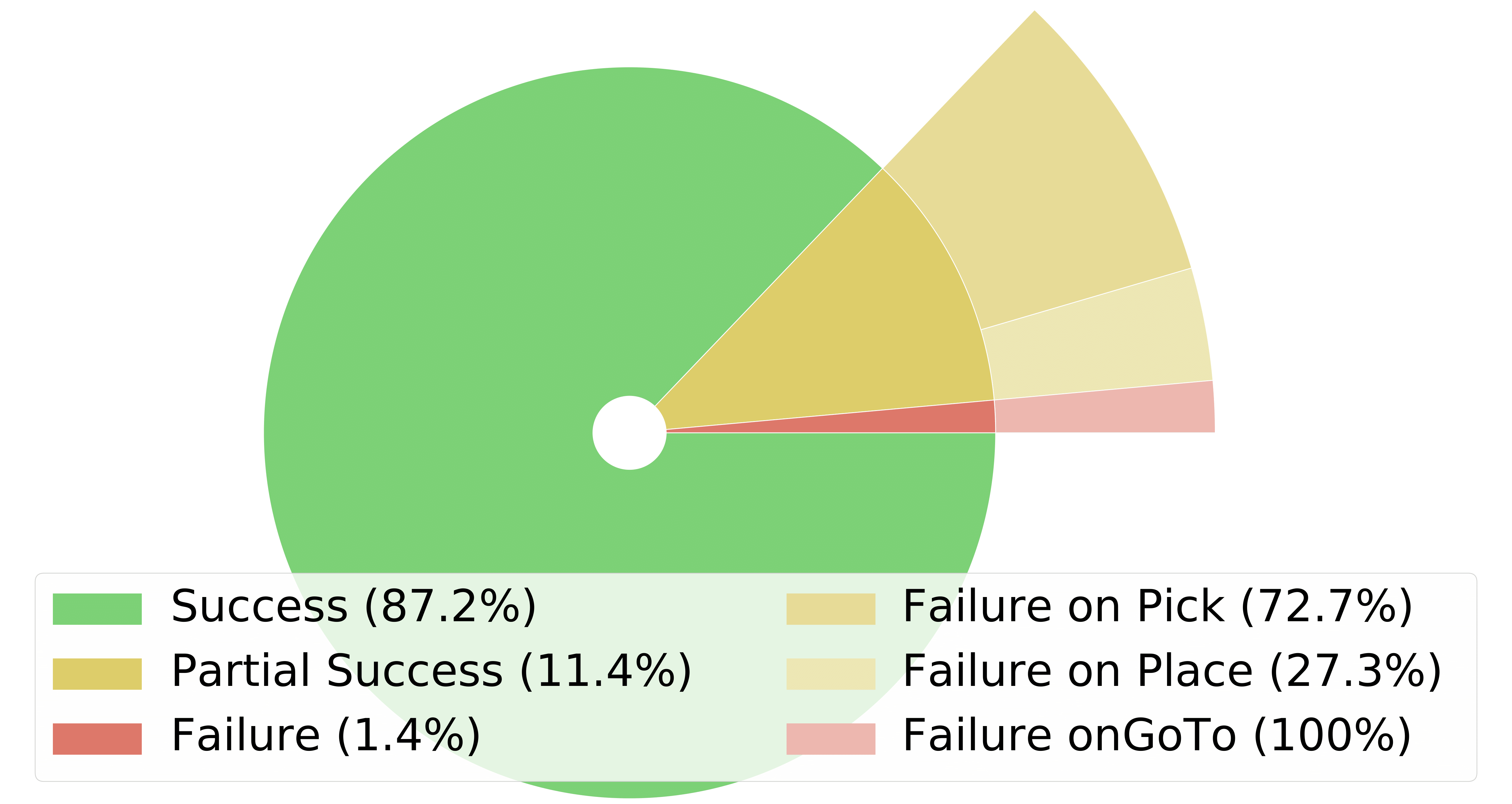}
        \caption{Task level.}
        \label{fig:task_success}
    \end{subfigure}
    \caption{Reliability analysis on $70$ tasks and $724$ objects.}
    \label{fig:reliability}
\end{figure}

We performed $70$ stocking tasks, involving a total of $724$ individual stocking operations (approximately $N=10$ objects per task on average). The results shown in Fig.~\ref{fig:reliability} demonstrate the robustness of our system, achieving an object-level success rate exceeding $98\%$. Most failures occurred during the picking stage. These failures typically did not halt task execution, though they led to partially completed tasks, in which not all objects in them were correctly placed.

At the task level, the success rate exceeded $87\%$. A detailed analysis revealed that out of the $9$ unsuccessful tasks, $8$ were partial failures, mainly due to minor placement inaccuracies linked to small ArUco detection errors (Sec.~\ref{sec:micro_navigation}). Only one task completely failed, necessitating human intervention due to total loss of the ArUco marker during the \textit{GoTo} action. Additionally, multiple pick failures occasionally appeared within a single task, which we believe to be related to network latency affecting image transmission and hence object classification.

When evaluating the performance of the perception pipeline, all picking failures were due to classification errors. Accuracy and recall metrics in Tab.~\ref{tab:recall} indicate better performance for fine-tuned YOLO models compared to GPT-4o-based counterparts. Crucially, all failures involved products lacking a dedicated YOLO model. These cases led to suboptimal segment selection and unrecoverable picking errors.

\begin{table}[b]
    \centering
    \begin{tabular}{c||c|c|c}
        \textbf{Model} &  \textbf{Accuracy} &\textbf{Recall} & \textbf{Inference}\\
        \hline\hline
        YOLO &  $97.5\%$ & $93.0\%$ & $0.2\si{s}$\\
        \hline
        GPT-4o &  $84.7\%$ & $82.6\%$ & $6.9\si{s}$\\
        \hline\hline
    \end{tabular}
    \caption{Comparison between YOLO and GPT-4o.}
    \label{tab:recall}
\end{table}

\subsection{Performance}
\begin{figure}[t]
    \centering    \includegraphics[width=\linewidth]{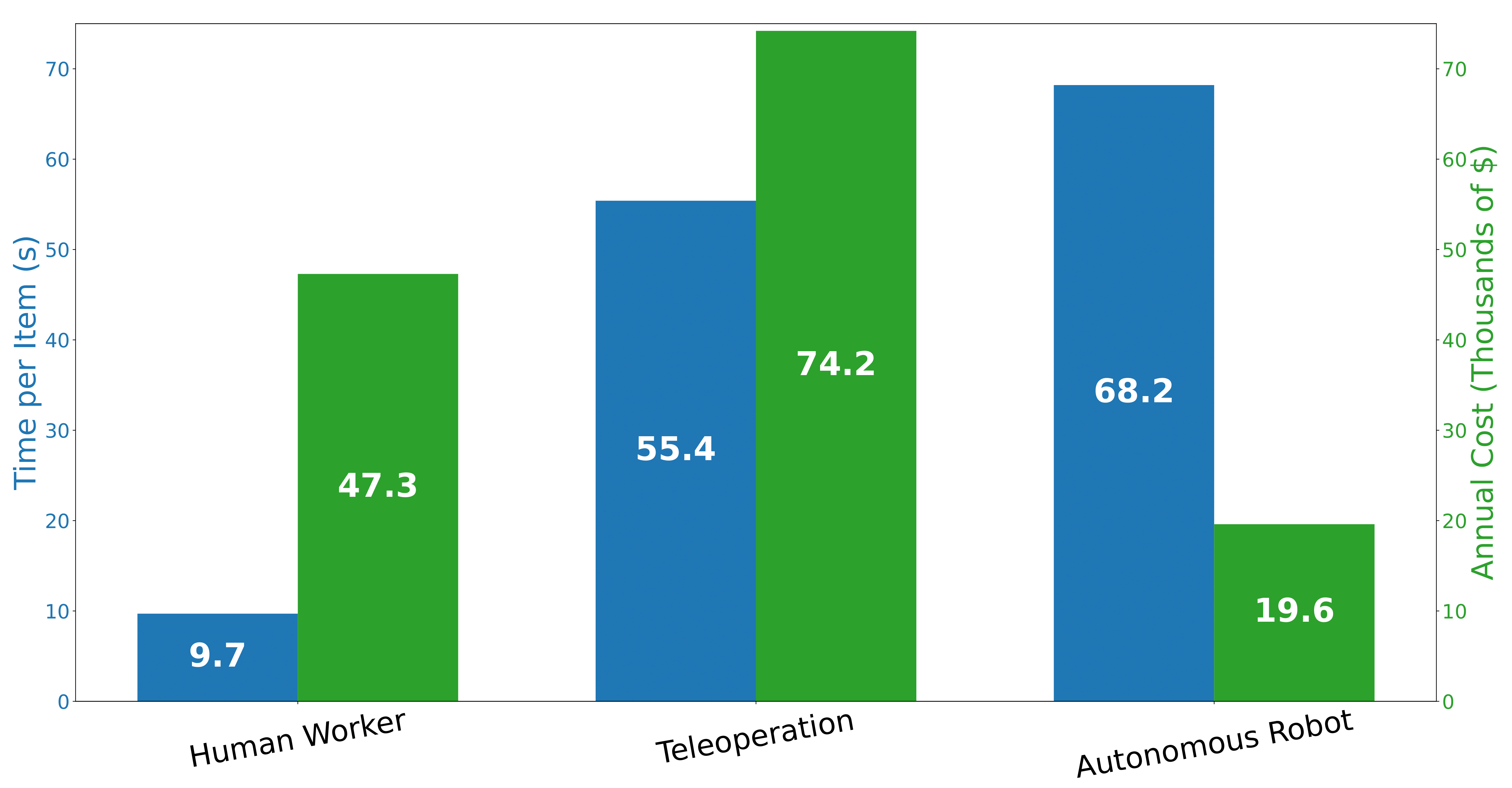}
    \caption{Comparison of cost and performance.}
    \label{fig:performces}
\end{figure}
We evaluated the autonomous stocking system's performance against teleoperation and manual human stocking in terms of stocking time and annual operational cost (Fig.~\ref{fig:performces}). Human stocking metrics were obtained from literature\cite{guney2019forecasting}, with labor cost derived from U.S. median salaries for these positions~\cite{bls_stockers_2024} and adjusted for non-wage employment cost~\cite{BLS2024ECEC}. Autonomous system yearly cost considered a \$25,000 robot amortized over a 5,000-hour lifespan, assuming 8-hour daily shifts with 2 hours of weekly maintenance. Teleoperation benchmarks included human operator salary and robotic costs, and performances after operator training.

The autonomous system was slower than teleoperation, largely due to sequential task execution and the lack of real-time cognitive adaptability available to human operators. Integrating advanced models like LLM showed impractical real-time performance, although using task-specific YOLO-based models improved picking times by about $6\si{s}$ per item (Tab.~\ref{tab:recall}), emphasizing the value of tailored solutions.

Both robotic methods still underperformed humans significantly, with manual labor approximately ten times faster. Cost analysis revealed teleoperation as economically nonviable, combining higher cost and subpar speed, whereas autonomous operation reduced costs considerably compared to humans, offering additional economic benefits such as overnight operation without incremental costs.

To contextualize these numbers, we contrast them with recent end-to-end retail mobile manipulation. \cite{airlab_paper} report in-store item picking trials with overall task success around $85\%$ and picking time of $50\si{s}$~\cite[ Sec. VIII.D]{airlab_paper}, excluding any navigation or placing of the item in a shelf. The system in \cite{in_the_wild} demonstrates autonomous picking in uncluttered environments with a $90\%$ success rate~\cite[Sec. VI.C]{in_the_wild} using a dual-arm platform. Instead, our framework achieves $98.3\%$ object-level success, including pick and place operations and shelf navigation, with a mean $68.2\si{s}$ per item for the full pipeline, yielding higher per-pick reliability and approximately 2 to 3 times faster pick execution.

Lastly, to quantify solution effectiveness in terms of combined time and cost, we defined a joint performance index:
\begin{equation}
p_i = \frac{1000}{\text{Annual Cost}\cdot\text{Time Per Item}},
\end{equation}
with higher values representing superior performance. By this metric, autonomous stocking ($p_{i,AR}=0.75$) performs worse than human workers ($p_{i,H}=2.18$) but surpasses teleoperation ($p_{i,T}=0.24$). Finally, a simple time decomposition using the times in Fig.~\ref{fig:performces} suggests that $\approx80\%$ of the gap can be attributed to hardware constraints and $\approx20\%$ to software limitations, highlighting hardware as the main focus of improvement towards human-level stocking efficiency.


\section{Conclusion}
\label{sec:conclusion}
We presented an autonomous stocking system demonstrating practical viability in supermarket environments, emphasizing cost-effectiveness and reliability through the integration of BTs for planning, improved ArUco-KF-based localization, a two-step MPC for precise navigation, and a robust visual perception pipeline based on fine-tuned YOLO models, SAM2 segmentation, and geometry-aware mask selection. Our approach was validated with extensive experiments and detailed analysis, highlighting both system robustness and key directions for future improvement.

Although robotic performance currently lags behind human efficiency due to hardware and software limitations (ArUco dependence, sparse packing, side grasps), the integration of advanced perception and planning algorithms improves operational robustness. Future enhancements to hardware capabilities and algorithmic optimizations promise significant steps toward deployment in retail automation.

\bibliography{biblio}
\bibliographystyle{IEEEtran}

\end{document}